# A Realistic and Robust Model for Chinese Word Segmentation

Chu-Ren Huang
Institute of Linguistics, Academia Sinica
Hong Kong Polytechnic University
churenhuang@gmail.com

Ting-Shuo Yo
TIGP CLCLP
Academia Sinica
tingshuo.yo@gmail.com

Petr Šimon
TIGP CLCLP
Academia Sinica
petr.simon@gmail.com

Shu-Kai Hsieh
Department of English
National Taiwan Normal University
shukai@gmail.com

## Abstract

A realistic Chinese word segmentation tool must adapt to textual variations with minimal training input and yet robust enough to yield reliable segmentation result for all variants. Various lexicon-driven approaches to Chinese segmentation, e.g. [1,16], achieve high f-scores yet require massive training for any variation. Text-driven approach, e.g. [12], can be easily adapted for domain and genre changes yet has difficulty matching the high f-scores of the lexicon-driven approaches. In this paper, we refine and implement an innovative text-driven word boundary decision (WBD) segmentation model proposed in [15]. The WBD model treats word segmentation simply and efficiently as a binary decision on whether to realize the natural textual break between two adjacent characters as a word boundary. The WBD model allows simple and quick training data preparation converting characters as contextual vectors for learning the word boundary decision. Machine learning experiments with four different classifiers show that training with 1,000 vectors and 1 million vectors achieve comparable and reliable results. In addition, when applied to SigHAN Bakeoff 3 competition data, the WBD model produces OOV recall rates that are higher than all published results. Unlike all previous work, our OOV recall rate is comparable to our own F-score. Both experiments support the claim that the WBD model is a realistic model for Chinese word segmentation as it can be easily adapted for new variants with robust result. In conclusion, we will discuss linguistic ramifications as well as future implications for the WBD approach.

## 1. Background and Motivation

The paper deals with the fundamental issue why Chinese word segmentation remains a research topic and not a language technology application after more than twenty years of intensive study. Chinese text is typically presented as a continuous string of characters without conventionalized demarcation of word boundaries. Hence tokenization of words, commonly called word segmentation in literature, is a pre-requisite first step for Chinese language processing. Recent advances in Chinese word segmentation (CWS) include popular standardized competitions run by ACL SigHAN and typically high F-scores around 0.95 from leading teams [8]. However, these results are achieved at the cost of high computational demands, including massive resources and long machine learning time. In fact, all leading systems are expected to under-perform substantially without prior substantial training. It is also important to note that SigHAN competitions are conducted under the assumption that a segmentation program must be tuned separately for different source texts and will perform differently. This is a bow to the fact that different communities may conventionalize the concept of word differently; but also an implicit concession that it is hard for existing segmentation programs to deal with textual variations robustly.

[15] proposed an innovative model for Chinese word segmentation which formulates it as simple two class classification task without having to refer to massive lexical knowledge base. We refine and implement this Word Boundary Decision (WBD) model and show that it is indeed realistic and robust. With drastically smaller demand on computational resources, we achieved comparable F-score with leading Bakeoff3 teams and outperform all on OOV recall, the most reliable criterion to show that our system deals with new events effectively.

In what follows, we will discuss modeling issues and survey previous work in the first section. The WBD model will be introduced in the second section. This is followed by a description of the machine learning model is trained in Section 4. Results of applying this implementation to SigHAN Bakeoff3 data is presented in Section 5. We conclude with discussion of theoretical ramifications and implications in Section 6.

## 2. How to model Chinese word segmentation

The performance of Chinese word segmentation (CWS) systems is directly influenced by their design criteria and how Chinese word segmentation task is modeled. These modeling issues did not receive in-depth discussion in previous literature:

**Modeling segmentation**. The input to Chinese word segmentation is a string of characters. However, the task of segmentation can be modeled differently. All previous work share the assumption that the task of segmentation is to find out all segments of the string that are words. This can be done intuitively by dictionary lookup, or by looking at strength of collocation within a string, e.g. [12]. Recent studies, e.g. [14, 16, 5, 17], reduce the complexity of this model and avoided the thorny issue of the elusive concept of word at the same time by modeling segmentation as learning the likelihood of characters being the edges of these word strings. These studies showed that, with sufficient features, machine learning algorithms can learn from training corpus and use their inherent model to tokenize Chinese text satisfactorily. The antagonistic null hypothesis of treating segmentation as simply identifying inherent textual breaks between two adjacent characters was never pursued.

**Out-of-Vocabulary words**. Identification of Out-of Vocabulary words (OOV, sometimes conveniently referred to as new words) has been a challenge to all systems due to data sparseness problem, as well as for dealing with true neologisms which cannot be learned from training data per se. This requirement means that CWS system design must incorporate explicit or implicit morphology knowledge to assure appropriate sensitivity to context in which potential words occur as previously unseen character sequences.

**Language variations,** especially among different Chinese speaking communities. Note that different Chinese speaking communities in PRC, Taiwan, Hong Kong, Singapore etc. developed different textual conventions as well as lexical items. This is compounded by the usual text type, domain, and genre contrasts. A robust CWS system must be able to adapt to these variations without requiring massive retraining. A production environment with it's time restrictions possesses great demands on the segmentation system to be able to quickly accommodate even to mixture of text types, since such a mixture would introduce confusing contexts and confuse system that would rely too heavily on text type, i.e. particular lexicon choice and specific morphology, and too large a context.

**Space and time demands**. Current CWS systems cannot avoid long training times and large memory demands. This is a consequence of the segmentation model employed. This is acceptable when CWS systems are used for offline tasks such as corpora preprocessing, where time and space can be easily provided and when needed. However, for any typically web-based practical language engineering applications, such high demand on computing time is not acceptable.

## 2.1 Previous works: a critical review

Two contrasting approaches to Chinese word segmentation summarize the dilemma of segmentation system design. A priori, one can argue that segmentation is the essential tool for building a (mental) lexicon hence segmentation cannot presuppose lexical knowledge. On the other hand, as a practical language technology issue, one can also argue that segmentation is simply matching all possible words from a (hypothetical) universal lexicon and can be simplified as mapping to a large yet incomplete lexicon. Hence we can largely divide previous approaches to Chinese word segmentation as lexicon-driven or text-driven.

**Text-Driven**. Text-driven approach to segmentation relies on contextual information to identify words and do not assume any prior lexical knowledge. Researches in this approach typically emphasize the need for an empirical approach to define the concept of a word in a language [12]. Work based on mutual information (MI) is the best-known and most comprehensive in this approach. The advantage of this approach can be applied to all different variations of language and yet be highly adaptive. However, the basic implementation of MI applies bi-syllabic words only. In addition, it cannot differentiate between highly collocative bigrams (such as 就不 *jiubu* "…then not…") and words. Hence it typically has lower recall and precision rate than current methods. Even though text-driven approaches are no longer popular, they are still widely used to deal with OOV with a lexicon-driven approach.

**Tokenization**. The classical lexicon-driven segmentation model, described in [1] is still adopted in many recent works. Segmentation is typically divided into two stages: dictionary look up and OOV word identification. This approach requires comparing and matching tens of thousands of dictionary entries in addition to guessing a good number of OOV words. In other words, it has a $10^4$ x $10^4$ scale mapping problem with unavoidable data sparseness. This model also has the unavoidable problem of overlapping ambiguity where e.g. a string [$C_{i-1}$, $C_i$, $C_{i+1}$] contains multiple sub-strings, such as [$C_{i-1}$, $C_i$] and [$C_i$, $C_{i+1}$], which are entries in the dictionary. The degree of such ambiguities is estimated to fall between 5% to 20% [2, 6].

**Character classification**. Character classification or tagging, first proposed in [14], became a very popular approach recently since it is proved to be very effective in addressing problems of scalability and data sparseness [14, 4, 16, 17]. Since it tries to model the possible position of a character in a word as character-strings, it is still lexicon-driven. This approach has been also successfully applied by name entity resolution, e.g. [17]. This approach is closely related to the adoption of the machine learning algorithm like conditional random field (CRF), [7]. CRF has been shown [11] to be optimal algorithm for sequence classification. The major disadvantages are big memory and computational time requirement.

## 3. Model

Our approach is based on a simplified idea of Chinese text, which we have introduced earlier in [15]. Chinese text can be formalized as a sequence of characters and intervals as illustrated in Figure 1.

$$c_1, I_1, c_2, I_2, ..., c_{n-1}, I_{n-1}, c_n$$

*Figure 1: Chinese text formalization*

There is no indication of word boundaries in Chinese text, only string of characters $c_i$. Characters in this string can be conceived as being separated by interval $I_i$. To obtain a segmented text, i.e. a text where individual words are delimited by some graphical mark such as space, we need to identify which of these intervals are to be replaced by such word delimiter.

We can introduce a utility notion of imaginary intervals between characters, which we formally classify into two types:

**Type 0**: a character boundary (CB) is an imaginary boundary between two characters

**Type 1**: a word boundary (WB), an interval separating two words.

With such a formulation, segmentation task can be easily defined as a classification task and machine learning algorithms can be employed to solve it. For conventional machine learning algorithms, classifications are made based on a set of features, which identify certain properties of the target to be classified.

In a segmented text, all the intervals between characters are labeled as a word boundary or as a character boundary, however, characters are not considered as being part of any particular word. Their sole function is to act as a contextual aid for identification of the most probable interval label. Since the intervals between characters (be it a word boundary or a character boundary) don't carry any information at all, we need to rely on the information provided by group of characters surrounding them.

Now we can collect n-grams that will provide data for construction of features that will provide learning basis for machine learning algorithm. A sequence, such the one illustrated in Figure 1, can be obtained from segmented corpus, and hence the probability of word boundary with specified relation to each n-gram may be derived. The resulting table which consists of each distinct n-gram entry observed in the corpus and the probability of a word boundary defines our n-gram collection.

Figure 2 shows the format of the feature vectors, or interval vectors, used in this study. We build the n-gram model up to n = 2.

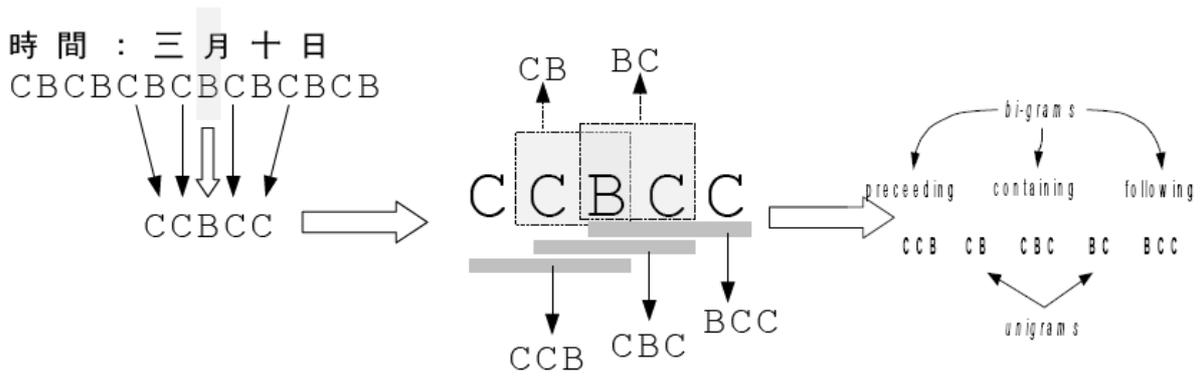

*Figure 2: The feature vectors used in this study. While C denotes a character in the sequence, B indicates the imaginary boundary. Thus CBC denotes a bi-gram containing the interval.*

To allow for more fine-grained statistical information we take into consideration five possible contributing surrounding contexts for each interval: two unigrams and three bi-grams. For convenience, we can define each interval by the two characters that surround it. Then, for each interval *<b,c>* in a 4-character context *abcd* we collect two unigrams *b* and *c* and three bi-grams *ab, bc,* and *cd* and compute probability of that interval being a word boundary given that particular context. These five n-grams are stored in a vector, which is labeled as Type 0 (character boundary) or Type 1 (word boundary) depending on the actual classification of that interval in the corpus: *<ab, b, bc, c, cd*, 0> or *<ab, b, bc, c, cd*, 1>. An example of an encoding of a sample from the beginning of Bakeoff 3 AS training corpus: "時間：三月十日" (shijian:sanyueshiri), which would be correctly segmented as "時間 ： 三月 十日" (shijian : sanyue shiri) can be seen in Table 1.

Set of such interval vectors provides a training corpus on which we apply machine learning algorithm, in our case logarithmic regression. Unsegmented text is prepared in the same fashion and the interval vectors are subsequently labeled by a classifier.

## 4. Training the Machine Learning Model

| ab | b | bc | c | cd | Typ. | Inter. |
|---|---|---|---|---|---|---|
| 0.500 | 0.595 | 0.003 | 0.173 | 0.021 | 0 | 時間 |
| 0.983 | 0.958 | 1.000 | 0.998 | 1.000 | 1 | 間： |
| 1.000 | 0.998 | 1.000 | 0.713 | 0.994 | 1 | ：三 |
| 0.301 | 0.539 | 0.010 | 0.318 | 0.054 | 0 | 三月 |
| 0.964 | 0.852 | 1.000 | 0.426 | 0.468 | 1 | 月十 |
| 0.002 | 0.245 | 0.065 | 0.490 | 0.010 | 0 | 十日 |

Table 1: Example of encoding and labeling of interval vectors in a 4-character window *abcd*

It is our goal to develop a segmentation system that would be able to handle different types of texts. A large uniform training corpus is desirable for high precision of segmentation, but that would cause a specialization of the classifier to types of texts contained in the corpus and system's generality would be compromised.

Furthermore, using a training data set converted from an independent corpus may give supplementary information and provide certain adaptation mechanism for the classifier during training, but leave the basic n-gram collection untouched. However, a smaller set of training data may give similar performance but with much lower cost.

If the features in the n-gram collection are properly defined, the final results from different machine learning algorithms may not differ too much. On the contrary, if the available n-gram collection does not provide efficient information, classifiers with ability to adjust the feature space may be necessary.

In our preliminary tests, during which we wanted to decide which machine learning algorithm would be most appropriate, the Academia Sinica Balance Corpus (ASBC) is used for the derivation of the n-gram collection and training data. The CityU corpus from the SigHAN Bakeoff2 collection is used for testing.

In order to verify the effect of the size of the training data, the full ASBC (~17 million intervals) and a subset of it (1 million randomly selected intervals) are used for training separately. Furthermore, four different classifiers, i.e., logistic regression (LogReg) [9], linear discriminative analysis (LDA) [13], multi-layer perceptron (NNET) [13], and support vector machine (SVM) [3], were tested.

The segmentation results are compared with the "gold standard" provided by the SigHAN Bakeoff2. Tables 2 and 3 show the training and testing accuracies of various classifiers trained with the ASBC. All classifiers tested perform as expected, with their training errors increase with the size of the training data, and the testing errors decrease with it. Table 2 clearly shows that the training data size has little effect on the testing error while it is above 1000. This proves that once a sufficient n-gram collection is provided for preparation of the interval vectors, classifier can be trained with little input.

| No of vectors | LogReg | LDA | NNet | SVM |
|---:|---:|---:|---:|---:|
| 17,577,301 | 0.9857 | 0.9784 | 0.9865 | 0.9862 |
| 1,000,000 | 0.9862 | 0.9796 | 0.9881 | 0.9876 |
| 100,000 | 0.9856 | 0.9796 | 0.9844 | 0.9867 |
| 10,000 | 0.9872 | 0.9811 | 0.9892 | 0.9879 |
| 1,000 | 0.9910 | 0.9820 | 0.9940 | 0.9920 |
| 100 | 1.0000 | 0.9700 | 1.0000 | 0.9900 |

*Table 2: Performance during training: corpus data from ASBC*

| No of vectors | LogReg | LDA | NNet | SVM |
|---:|---:|---:|---:|---:|
| 17,577,301 | 0.9386 | 0.9326 | 0.9373 | 0.9362 |
| 1,000,000 | 0.9386 | 0.9325 | 0.9360 | 0.9359 |
| 100,000 | 0.9389 | 0.9326 | 0.9331 | 0.9369 |
| 10,000 | 0.9393 | 0.9326 | 0.9338 | 0.9364 |
| 1,000 | 0.9373 | 0.9330 | 0.9334 | 0.9366 |
| 100 | 0.9106 | 0.9355 | 0.9198 | 0.9386 |

*Table 3: Performance during testing: corpus data from SigHAN BakeOff2*

It is also shown in Table 2 that four classifiers give similar performance when the training data size is above 1000. However, while the training sample size drops to 100, the SVM and LDA algorithms show their strength by giving similar performance to the experiments trained with larger training data sets.

To further explore the effectiveness of our approach, we have modified the experiment to show the performance in model adaptation. In the modified experiments the training and testing data sets are both taken from a foreign corpus (CityU), while our n-gram collection is still from ASBC. The relation between the derived features and the true segmentation may be different from the ASBC, and hence is learned by the classifiers. The results of the modified experiments are shown in Tables 4 and 5.

| No of vectors | LogReg | LDA | NNet | SVM |
|---:|---:|---:|---:|---:|
| 1,000,000 | 0.9477 | 0.9458 | 0.9495 | 0.9500 |
| 100,000 | 0.9484 | 0.9461 | 0.9496 | 0.9504 |
| 10,000 | 0.9491 | 0.9470 | 0.9510 | 0.9525 |
| 1,000 | 0.9460 | 0.9440 | 0.9520 | 0.9530 |
| 100 | 0.9600 | 0.9600 | 0.9700 | 0.9700 |

*Table 4: Performance during training: new corpus data from cityU*

| No of vectors | LogReg | LDA | NNet | SVM |
|---:|---:|---:|---:|---:|
| 1,000,000 | 0.9424 | 0.9390 | 0.9423 | 0.9443 |
| 100,000 | 0.9425 | 0.9387 | 0.9417 | 0.9441 |
| 10,000 | 0.9421 | 0.9410 | 0.9409 | 0.9430 |
| 1,000 | 0.9419 | 0.9418 | 0.9332 | 0.9400 |
| 100 | 0.8857 | 0.9350 | 0.8812 | 0.9299 |

*Table 5: Performance during testing: new corpus data from cityU*

## 5. Results

In our test to compare our performance objectively with other approaches, we adopt logarithmic regression as our learning algorithm as it yielded best results during our test. We apply the segmentation system to two traditional Chinese corpora, CKIP and CityU, provided for SigHAN Bakeoff 3. In the first set of tests, we used training corpora provided by SigHAN Bakeoff3 for n-gram collection, training and testing. Results of these tests are presented in Table 6.

In addition, to underline the adaptability of this approach, we also tried combining both corpora and then ran training on random sample of vectors. This set of tests is designed to exclude the possibility of over-fitting and to underline the robustness of the WBD model.

Note that such tests are not performed in SigHAN Bakeoffs as many of the best performances are likely over-fitted. Results of this test are shown in Table 7.

|                | cityu | ckip  |
|----------------|-------|-------|
| F-measure      | 0.933 | 0.919 |
| OOV Rate       | 0.179 | 0.204 |
| OOV Recall Rate| 0.888 | 0.871 |
| IV Recall Rate | 0.941 | 0.943 |

Table 6: Results (Bakeoff 3 dataset): traditional Chinese

|                | cityu | ckip  |
|----------------|-------|-------|
| F-measure      | 0.920 | 0.925 |
| OOV Rate       | 0.167 | 0.187 |
| OOV Recall Rate| 0.920 | 0.893 |
| IV Recall Rate | 0.920 | 0.930 |

Table 7: Combined results (Bakeoff 3 dataset): traditional Chinese

Table 6 and 7 show that our OOV recall is comparable with our overall F-score, especially when our system is trained on selected vectors from combined corpus. This is in direct contrast with all existing systems, which typically has a much lower OOV recall than IV recall. In other words, our approach applies robustly to all textual variations with reliably good results. Table 8 shows that indeed our OOV recall rate shows over 16% improvement over the best Bakeoff3 result for CityU, and over 27% improvement over best result for CKIP data.

|                          | ckip  | cityu |
|--------------------------|-------|-------|
| Microsoft Research Asia  | 0.702 | 0.792 |
| IASL                     | 0.656 | 0.792 |
| Respective corpus        | **0.888** | **0.871** |
| Combined corpora         | **0.893** | **0.920** |

Table 6: Our OOV recall results compared to best performing systems in (Levow, 2006)

## 6. Discussion

We refined and implemented the WBD model for Chinese word segmentation and show that it is a robust and realistic model for Chinese language technology. Most crucially, we show that the WBD model is able to reconcile the two competitive goals of the lexicon-driven and text-driven approaches. The WBD model maintains comparable F-score level with the most recent CRF character-classification based results, yet improves substantially on the OOV recall.

We showed that our system is robust and not over-fitted to a particular corpus, as it yields comparable and reliable results for both OOV and IV words. In addition, we show that same level of consistently high results can be achieved across different text sources. Our results show that Chinese word segmentation system can be quite efficient even when using very simple model and simple set of features.

Our current system, which has not been optimized for speed, is able to segment text in less then 50 seconds. Time measurement includes preparation of testing data, but also training phase. We believe that with optimized and linked computing power, it will be easy to implement a real time application system based on our model. In the training stage, we have shown that sampling of around 1,000 vectors is enough to yield one of the best results. Again, this is a promised fact for the WBD model of segmentation to be robust. It is notable, that in case of training on combined corpora (CKIP and CityU) the results are even better than test in respective data sets, i.e. CKIP training corpus for segmenting CKIP testing text, or CityU respectively. This is undoubtedly the result of our strategy of granulation of the context around each interval. Since four characters that we use for representation of the interval context are broken up into two unigrams and three bi-grams, we let the system to get more refined insight into the segmented area.

Consequently, the system is learning morphology of Chinese with greater generality and this results in higher OOV scores. It can be argued that in our combined corpora test, the OOV recall is even higher, because the input contains two different variants of Chinese language, Taiwanese variant contained in CKIP corpus and Hong Kong variant contained in CityU corpus.

Text preparation and post-processing also add to overall processing time. In our current results, apart from context vector preparation there was no other preprocessing employed and neither any post-processing. This fact also shows that our system is able to handle any type of input without the need to define special rules to pre- or post-process the text. Early results applying our model to simplified Chinese corpora are also promising.

In sum, our WBD model for Chinese word segmentation yields one of the truly robust and realistic segmentation program for language technology applications. If these experiments are treated as simulation, our results also support the linguistic hypothesis that word can be reliably discovered without a built-in/innate lexicon. We will look into developing a more complete model to allow for more explanatory account for domain specific shifts as well as for effective bootstrapping with some lexical seeds.